\documentclass[a4wide]{article}

\usepackage[margin=1.5in]{geometry}
\usepackage{amssymb, amsmath, amsopn}
\usepackage{times,mathptmx}
 \usepackage{mathptmx}
 \usepackage{helvet}
 \usepackage{courier}
 \usepackage{makeidx}
 \usepackage{multicol}
 \usepackage{footmisc}
 \usepackage[justification=centering]{caption}
 \usepackage{amssymb, amsmath}
 \usepackage{graphicx}
 \usepackage{caption}
 \usepackage{subfig}
 \usepackage{xspace}
 \usepackage{color}
 \usepackage{array}
 \newcolumntype{P}[1]{>{\centering\arraybackslash}p{#1}}
 \usepackage{verbatim}
 \usepackage{imakeidx, epsfig,url}
\usepackage{algorithm}
\usepackage[algo2e,ruled,vlined]{algorithm2e}
\usepackage{xspace}
\usepackage[numbers,sort&compress,longnamesfirst,sectionbib]{natbib}
\usepackage{balance}

\usepackage{graphicx}
\usepackage{rotating}
\usepackage{amssymb}
\usepackage{graphicx}
\usepackage{url} 

\usepackage{color}
\usepackage{xspace}
\usepackage{blindtext}
\usepackage{multirow, rotating}
\usepackage{algorithm,algpseudocode,booktabs,amsmath}
\makeatletter
\renewcommand{\ALG@beginalgorithmic}{\small}
\makeatother

\usepackage{hyperref}
\usepackage{balance}
\usepackage{tikz}
\usetikzlibrary{shapes,decorations,patterns,positioning}
\usetikzlibrary{plotmarks}
\graphicspath{{/plots/}{plots/}}
\newcommand{\ignore}[1]{}

\newcommand{\ea}{(1+1)~EA\xspace}

\usepackage{url}

\begin{document}

\title{The Evolutionary Process of Image Transition in Conjunction with Box and Strip Mutation}

\author{Aneta Neumann, Bradley Alexander, Frank Neumann}


\maketitle

\begin{abstract}
Evolutionary algorithms have been used in many ways to generate digital art.
We study how evolutionary processes are used for evolutionary art and present a new approach to the transition of images.  
Our main idea is to define evolutionary processes for digital image transition, combining different variants of mutation and evolutionary mechanisms. 
We introduce box and strip mutation operators which are specifically designed for image transition. Our experimental results show that the process of an evolutionary algorithm in combination with these mutation operators can be used as a valuable way to produce unique generative art.

\end{abstract}

\sloppy

\section{Introduction}\label{sec:intro}


The field of evolutionary algorithms (EAs) has been successfully applied to the areas of modern  art~\cite{bentley2002creative,citeulike:12541313}.
Evolutionary computation is an interesting approach to the generation of novel images and is beginning to have a broader impact on artistic fields. More generally, evolutionary methods applied to problem solving in creative fields is an exciting and fast developing topic in computer science.
In earlier years research has been conducted using evolutionary algorithms for interactive generation of art.  Dawkins~\cite{dawkins1986blind} and Smith~\cite{smith1982evolution} demonstrated the potential of Darwinian variation to evolve biomorphs graphic objects. Following this steps Sims~\cite{DBLP:conf/siggraph/Sims91}, Latham and Todd~\cite{Todd:1994:EAC:561831} have combined evolutionary techniques and computer graphics to create artistic images and to reproduce computer creatures of great complexity. More recently Unemi~\cite{rooke}, Hart ~\cite{DBLP:conf/siggraph/Hart06} created a collection of images using refined means of combining individual images and their genotypes.
These works have not only influenced the field of evolutionary computation, but were also displayed in galleries and museums around the world. Furthermore, research in computational art has employed deep neural networks to create artistic images~\cite{DBLP:journals/corr/Champandard16,DBLP:journals/corr/LiW16}.

In our work we introduce a novel approach to define evolutionary processes for evolutionary image transition~\footnote{Images and videos are available at \url{https://evolutionary-art.blogspot.com}}. The main idea comprises of using well-known evolutionary processes and adapting these in an artistic way to create innovative images.

\begin{figure}[t] 
\begin{center}  
\includegraphics[scale=0.5]{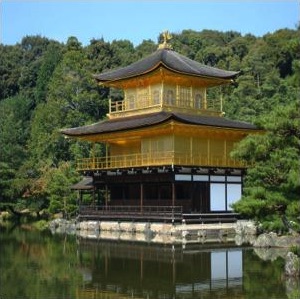} 
\includegraphics[scale=0.5]{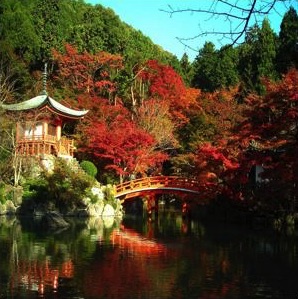} 

\end{center}   
\caption{Starting image $S$ (Kinkaku-Ji Temple Kyoto) and target image $T$ (Daigo-ji Temple Kyoto)}
\label{fig:opera}
\vspace{-0.3cm}
\end{figure}

The evolutionary transition process starts from a given starting image $S$ and, through a stochastic process, evolves it towards a target image $T$. Our goal is to maximise the fitness function where we count the number of the pixels matching those of the target image $T$. This problem mirrors the classical OneMax problem, which has been widely studied in academic literature~\cite{DBLP:journals/tcs/DrosteJW02,DBLP:journals/ec/JansenS10}.

We start by introducing a variant of the asymmetric mutation operator studied in the context of minimum spanning trees~\cite{DBLP:journals/tcs/NeumannW07} and pseudo-Boolean optimization\cite{DBLP:journals/ec/JansenS10}. This operator is based on flipping pixels with a given probability in each step and avoids the coupon collectors effect~\cite{Mitz2005} by using standard bit mutation applied to OneMax.
Afterwards, we introduce specific mutation operators for evolutionary image transition, namely strip and box mutations, which pick parts of the image consisting of strips and boxes and set them to the target image. Our experimental results show the artistic effects which can be produced by applying evolutionary process to an image via the operators described below.

The rest of the paper is structured as follows. In Section~\ref{sec2}, we introduce evolutionary transition process and examine the behaviour of~\ea for image transition in Section~\ref{sec3}.
Section~\ref{sec4} gives results of how \emph{strip mutation, combined strip mutation} and \emph{box mutation} can be used for evolutionary image transition process. In Section~\ref{sec5} we investigate the use of \emph{asymmetric mutation} as part of mutation operators and study their combinations with pixel-based mutations during the evolutionary process. Finally, we conclude with our overview about the insights from our investigations.


\section{Evolutionary Image Transition}
\label{sec2}

Artificial intelligence has the potential to change our perception of the process of creating art. Using evolutionary algorithms can lead to significant novelty in the  interaction between the computer and the artist in order to produce an artwork. The application of evolutionary algorithms in an innovative way can change the apprehension of human-machine creativity. Artificial intelligence as a medium for creating art has a long and well-established tradition. The growing influence and demand of artificial intelligence in our global industrial life has been discussed throughout a wide spectrum of our society. So comes a meaningful question: how can computational creativity perform as a valuable addition to digital art to produce unique generative art?

\begin{algorithm}[t]

\begin{itemize} 
\item Let $S$ be the starting image and $T$ be the target image.
\item Set X:=S.
\item Evaluate $f(X,T)$.
\item while (not termination condition)
\begin{itemize}
\item Obtain image $Y$ from $X$ by mutation.
\item Evaluate $f(Y, T)$
\item If $f(Y, T) \geq f(X, T)$, set $X:= Y$. 
\end{itemize}
\end{itemize}
\caption{Evolutionary algorithm for image transition}
\label{alg:ea}
\vspace{-0.3cm}
\end{algorithm}

We consider an evolutionary transition process that transforms a given image $X$ of size $m \times n$ into a given target image $T$ of size $m \times n$. Our goal is to study different ways of carrying out this evolutionary transformation based on random processes from an artistic perspective.
We start our process with a starting image $S=(S_{ij})$. Our algorithms evolve $S$ towards $T$ and have at each point in time an image $X$ where $X_{ij} \in \{S_{ij}, T_{ij} \}$. We say that pixel $X_{ij}$ is in state $s$ if $X_{ij} = S_{ij}$, and $X_{ij}$ is in state $t$ if $X_{ij}=T_{ij}$.
For our process we assume that $S_{ij} \not = T_{ij}$ as pixels with $S_{ij} = T_{ij}$ can not change values and therefore do not have to be considered in the evolutionary process.
To illustrate the effect of the different methods presented in this paper, we consider two images of famous Kyoto Temples (see Figure~\ref{fig:opera}).

The fitness function of an evolutionary algorithm guides its search process and determines how to move between images. Therefore, the fitness function itself has a strong influence on the artistic behaviour of the evolutionary image transition process. An important property for evolutionary image transition, in this work, is that images close to the target image get a higher fitness score. 
We measure the fitness of an image $X$ as the number of pixels where $X$ and $T$ agree. This fitness function is isomorphic to that of the OneMax problem when interpreting the pixels of $S$ as $0$'s and the pixels of $T$ as $1$'s.
Formally, we define the fitness of $X$ with respect to $T$ as

\ignore{
\[
f(X,T) = \sum_{i=1}^m \sum_{j=1}^n |x_{ij} - t_{ij}|
\]
}

\vspace{-0.2cm}
\[
f(X,T) = |\{X_{ij} \in X \mid X_{ij}=T_{ij}\}|.
\]
\vspace{-0.3cm}

We consider simple variants of the classical \ea in the context of image transition. The algorithm is using mutation only and accepts an offspring if it is at least as good as its parent according to the fitness function. The approach is given in Algorithm~\ref{alg:ea}.
Using this algorithm has the advantage that parents and offspring do not differ too much in terms of pixel which ensures a smooth process for transitioning the starting image towards the target.

All experimental results in this paper are shown for the process of moving from the starting image to the target image given in Figure~\ref{fig:opera} where the images are of size $200 \times 200$ pixels. The algorithms have been implemented in Matlab.
In order to visualize the process, we show the images obtained when the evolutionary process reaches 12.5\%, 37.5\%, 62.5\% and 87.5\% of pixels evolved towards the target image. 


\section{Asymmetric Mutation}
\label{sec3}


We consider a simple evolutionary algorithm that has been well studied in the area of runtime analysis, namely variants of the classical (1+1)~EA ~\cite{DBLP:journals/tcs/DrosteJW02}. As already mentioned, our setting for the image transition process is equivalent to the optimization process for the classical benchmark function OneMax. 
The standard variant of the \ea  flips each pixel with probability $1/|X|$ where $|X|$ is the total number of pixels in the given image. Using this mutation operator, the algorithm encounters the well-known coupon collector's effect which means that the additive drift towards the target image when having $k$ missing target pixels is $\Theta(k/n)$~\cite{DBLP:journals/ai/HeY01}.

\begin{figure}[t] 

\centering
\vskip -6pt

\includegraphics[scale=0.35]{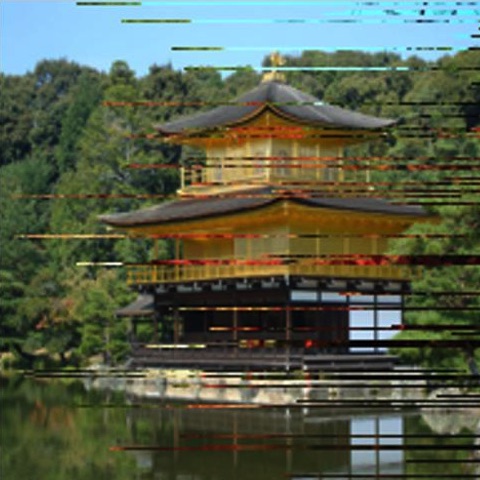} 
\includegraphics[scale=0.35]{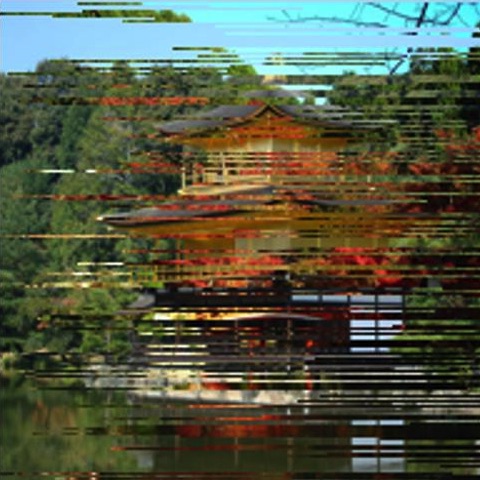} \hfil\\[-1.1\tabcolsep]
\vspace{0.3cm}
\includegraphics[scale=0.35]{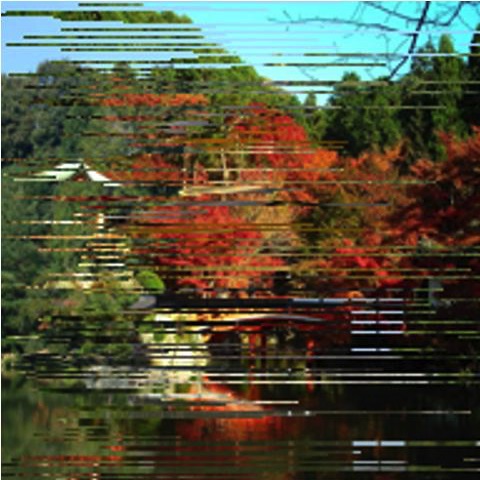}
\includegraphics[scale=0.335]{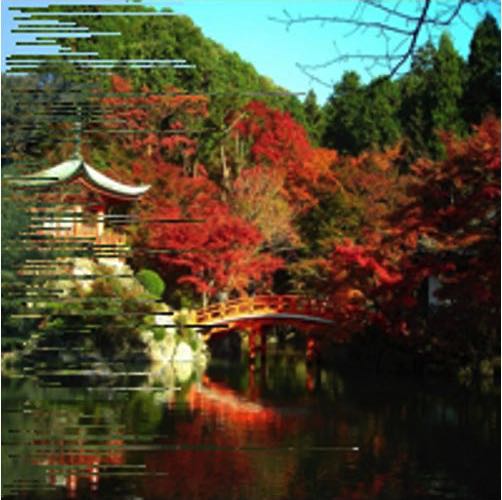}

\vspace{0.3cm}
\caption{Image transition for strip mutation}

\label{fig:2}
\end{figure}

\ignore{
\begin{algorithm}[t]
\vspace{-0.1cm}
\begin{itemize}
\item Obtain $Y$ from $X$ by flipping each pixel $X_{ij}$ with probability $c/|X|$ where $|X|$ is the number of pixels in $X$ and $c \geq 1$ is a constant.
\end{itemize}
\caption{Standard mutation}
\label{alg:mutstand}
\end{algorithm}
}
\begin{algorithm}[t]
\vspace{0.15cm}
\begin{itemize}
\item Obtain $Y$ from $X$ by flipping each pixel $X_{ij}$ of $X$ independently of the others with probability $c_s/(2|X|_S)$ if $X_{ij}=S_{ij}$, and flip $X_{ij}$ with probability $c_t/(2|X|_T)$ if $X_{ij}=T_{ij}$, where $c_s \geq 1$ and $c_t \geq 1$ are constants.
\end{itemize}
\vspace{-0.15cm}
\caption{Asymmetric mutation}
\label{alg:asym}
\end{algorithm}
In order to avoid the coupon collector's effect but obtaining images where all pixel are treated symmetrically, we use the
\emph{asymmetric mutation} operator introduced in \cite{DBLP:journals/tcs/NeumannW07}.
Jansen and Sudholt~\cite{DBLP:journals/ec/JansenS10} have shown that the (1+1)~EA using \emph{asymmetric mutation} optimizes OneMax in time $\Theta(n)$ which improves upon the usual bound of $\Theta(n \log n)$ when using standard bit mutations. 
We denote by $|X|_T$ the number of pixels where $X$ and $T$ agree. Similarly, we denote by $|X|_S$ the number of pixels where $X$ and $S$ agree. Each pixel is starting state $s$ is flipped with probability $c_s/(2|X|_S)$ and each pixel in target state $t$ is flipped with probability $c_t/(2|X|_T)$.
The mutation operator is shown in Algorithm~\ref{alg:asym}.
The mutation operator differs from the one given in~\cite{DBLP:journals/ec/JansenS10} by the two constants $c_s$ and $c_t$ which allow one to determine the expected number of new pixels from the starting image  and the target image, respectively. The choice of $c_s$ and $c_t$ determines the expected number of pixel in the starting state and target state to be flipped. To be precise, the expected number of pixels currently in the starting state $s$ to be flipped is $c_s/2$ and the number of pixels in the target state $t$ to be flipped is $c_t/2$ as long as the current solution $X$ contains at least that many pixel of the corresponding type.
 In \cite{DBLP:journals/ec/JansenS10} the case $c_s = c_t=1$ has been investigated which ensures that at each point in time has an additive drift of $\Theta(1)$ and therefore avoiding the coupon collector effect at the end of the process. Using different values for $c_s$ and $c_t$ allows us to change the speed of transition as well as the relation of the number of pixels switching from the starting image to the target and vice versa while still ensuring that there is constant progress towards the target. For all our experiments we use $c_s=100$ and $c_t=50$.

\section{Strip and Box Mutation}
\label{sec4}

\begin{figure}[t] 

\centering
\vskip -6pt

\includegraphics[scale=0.35]{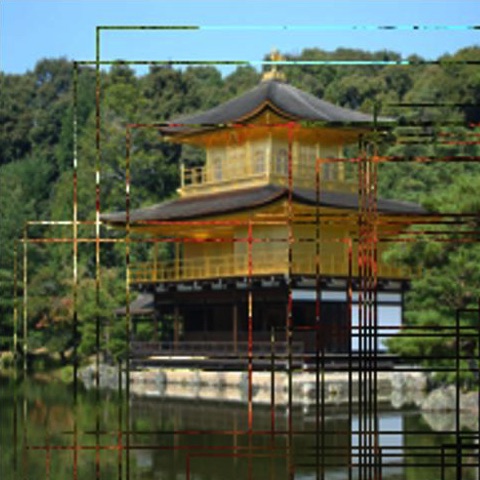}
\includegraphics[scale=0.35]{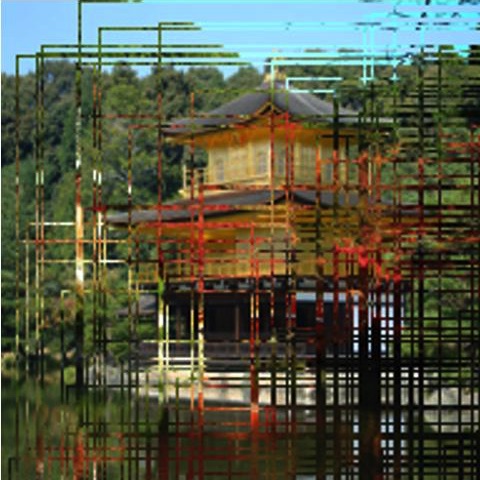} \hfil\\[-1.1\tabcolsep]
\vspace{0.3cm}
\includegraphics[scale=0.35]{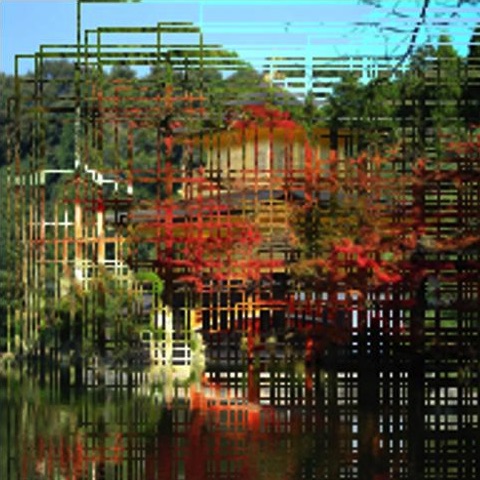}
\includegraphics[scale=0.3358]{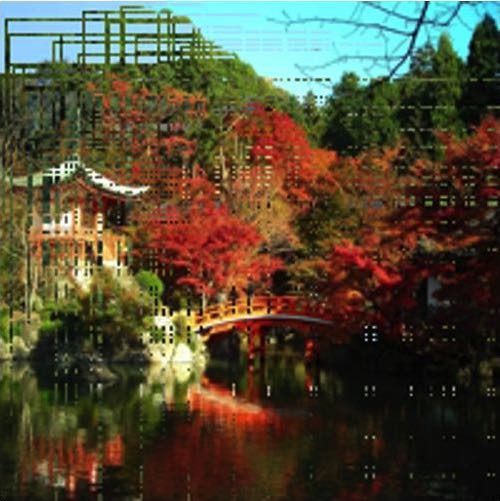} 

\vspace{0.3cm}
\caption{Image transition for combined strip mutation}

\label{fig:3}
\end{figure}


We now introduce specific mutation operators for image transition. The design of our three mutation operators, namely \emph{strip mutation}, \emph{combined strip mutation} and \emph{block mutation},  is strongly oriented towards production of interesting artistic images.

All of these mutation operators transition a region of the current image $X$ to an area of the target image $T$, starting at randomly chosen pixel position $X_{ij}$. \emph{Strip mutation} mutates a vertical strip of pixels from $X_{ij}$ for a length of $180$ pixels or to the boundary of the image whichever comes first. 
\emph{The strip mutation} imitates well-know technique in generative art called a glitch. This effect intentionally corrupts data of an image by encoding the JPEG process. All experiments shows that in each new generation the initial image and the target image overlay through a series of randomly placed strips. In Figure~\ref{fig:2} we show the experimental results of \emph{strip mutation}. The effect of \emph{strip mutation} creates generative art with highly interesting components. Furthermore, we have conceptualised a \emph{combined strip mutation} operator where both horizontal and vertical strips gradually overlay the original image in random locations. The parameter settings for \emph{the combined strip mutation} are $200$ pixels in width and $40$ pixels in height for the horizontal strip and one pixel in width and $200$ pixels in height for the vertical orientation.
Figure~\ref{fig:3} shows a very interesting transition from the point of view of an image processing which also has artistic value.

Our third mutation operator is \emph{block mutation}. This operator randomly selects a position in the space of the matrix $S (i,j)$ and flips a block of $3 \times 3$ pixels. Because the position chosen for mutation can be any point $X_{ij}$ we can observe that mutated blocks can overlap.

\begin{figure}[t] 

\centering
\vskip -6pt

\includegraphics[scale=0.35]{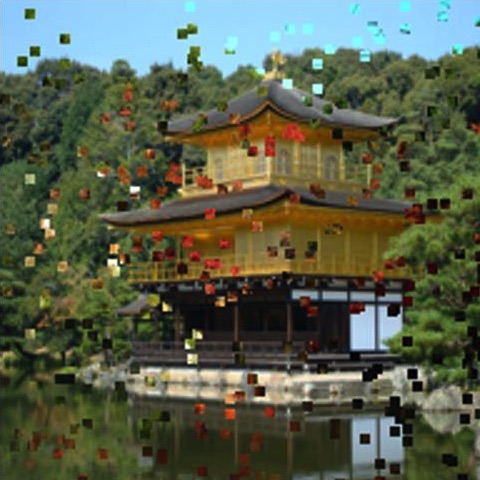}
\includegraphics[scale=0.35]{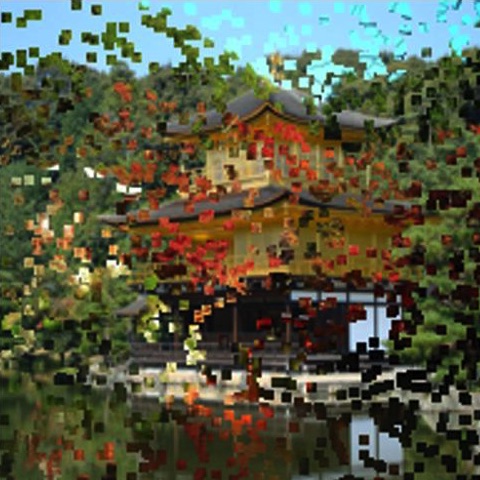} \hfil\\[-1.1\tabcolsep]
\vspace{0.3cm}
\includegraphics[scale=0.35]{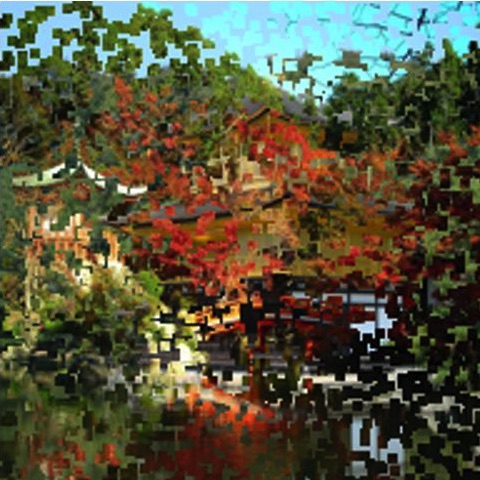}
\includegraphics[scale=0.335]{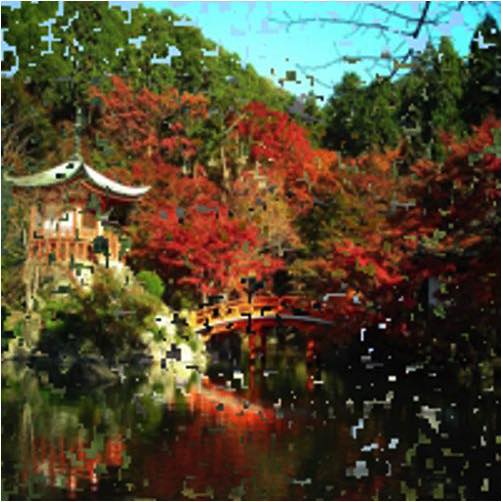} 

\vspace{0.3cm}
\caption{ Image transition for box mutation}

\label{fig:4}
\end{figure}

Figure~\ref{fig:4} shows the experimental results of \emph{block mutation}. We have executed a smooth transition so as to interest the viewer. Additionally, we have made the changes in the image clearly visible.

\section{Combined Approaches}
\label{sec5}


\emph{The asymmetric, strip, combined strip} and \emph{box mutation} operators have quite distinct behaviours when applied to image transition. We now study the effect of combining the approaches for evolutionary image transition in order to obtain a more artistic evolutionary process.
We explore the combination of \emph{the asymmetric mutation} operator and \emph{strip, combined strip} and \emph{box mutation}. Here we run \emph{the asymmetric mutation} operator as described in Algorithm~\ref{alg:asym}. The process alternates between \emph{asymmetric mutation} and one of either \emph{strip}, \emph{combined strip} or \emph{box mutation}.

\begin{figure}[t] 

\centering
\vskip -6pt

\includegraphics[scale=0.35]{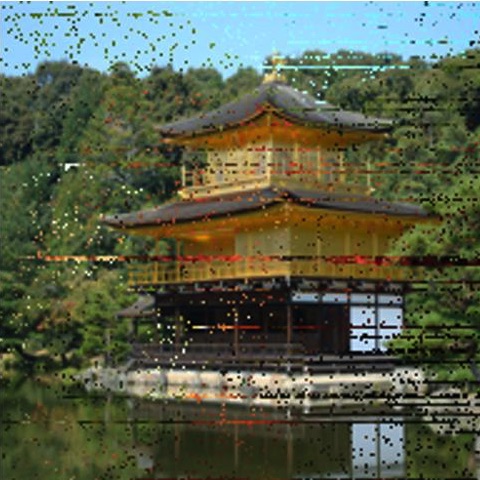}
\includegraphics[scale=0.35]{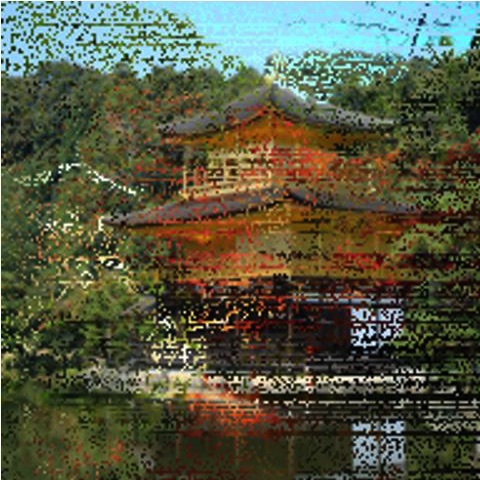} \hfil\\[-1.1\tabcolsep]
\vspace{0.3cm}
\includegraphics[scale=0.35]{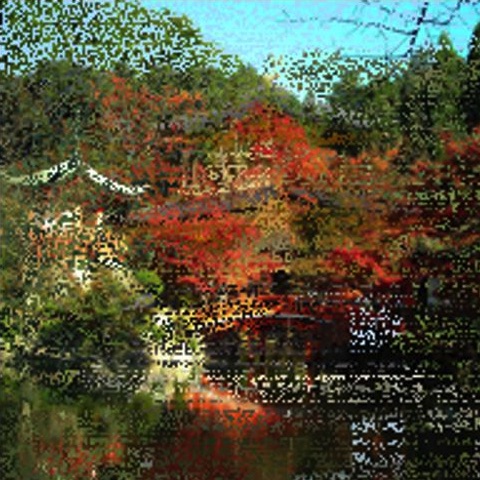}
\includegraphics[scale=0.3358]{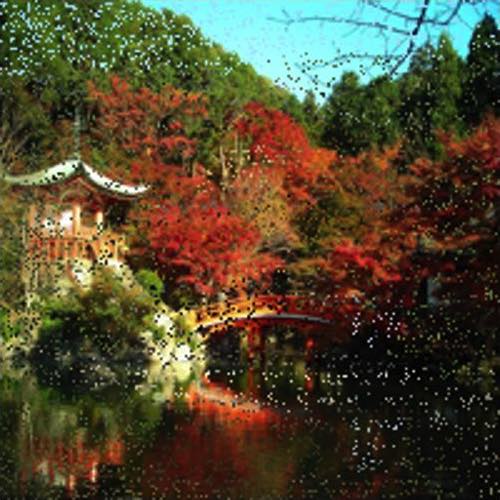} 

\vspace{0.3cm}
\caption{ Image transition for combined asymmetric and strip mutation}

\label{fig:5}
\end{figure}

Figure~\ref{fig:5} shows the results of the experiments for \emph{combined asymmetric} and \emph{strip mutation}. The transitional images are significantly different to the images produced in \emph{strip mutation} or \emph{box mutation}.

\begin{figure}[t] 

\centering
\vskip -6pt

\includegraphics[scale=0.35]{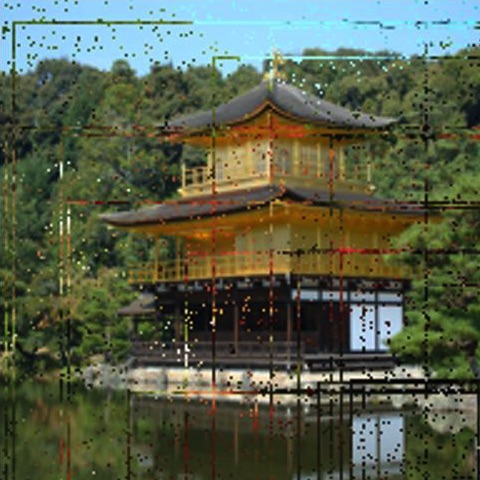}
\includegraphics[scale=0.35]{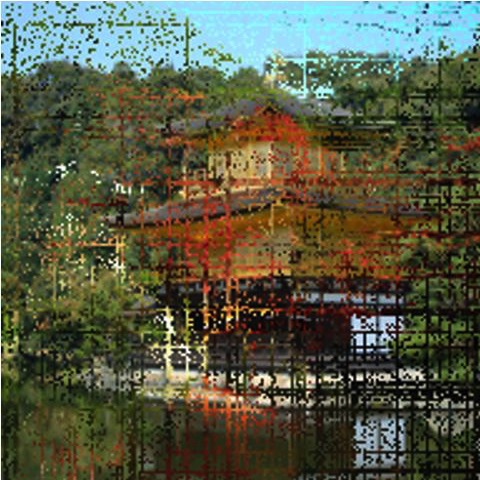} \hfil\\[-1.1\tabcolsep]
\vspace{0.3cm}
\includegraphics[scale=0.35]{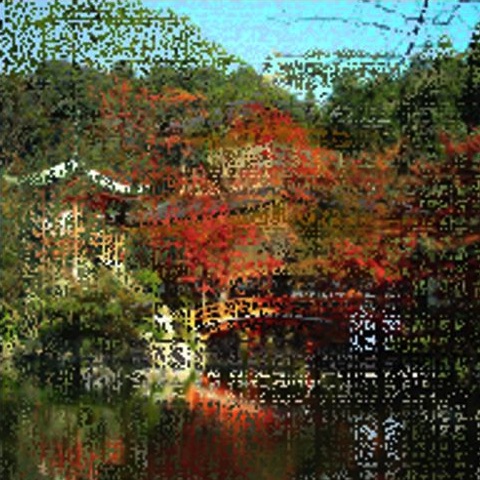}
\includegraphics[scale=0.335]{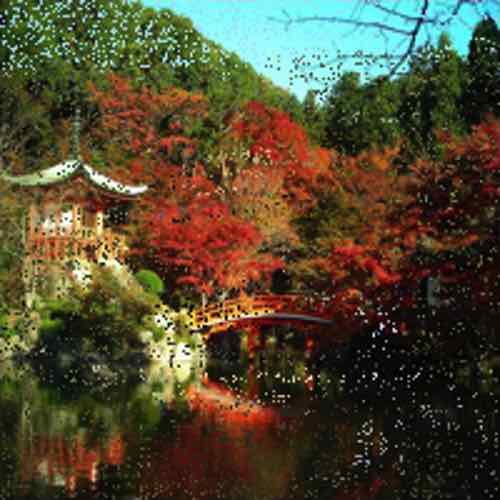} 

\vspace{0.3cm}
\caption{ Image transition for combined asymmetric and combined strip mutation}

\label{fig:7}
\end{figure}

\begin{figure}[t] 

\centering
\vskip -6pt

\includegraphics[scale=0.35]{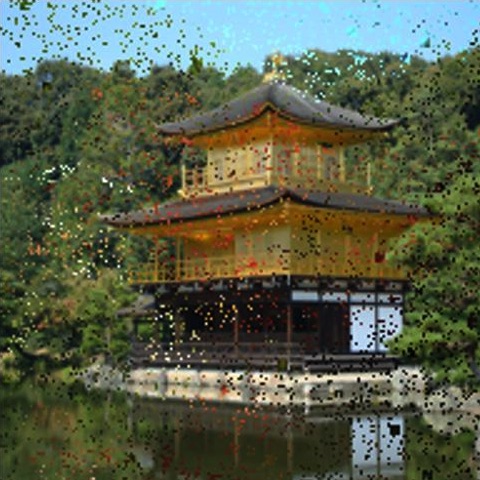}
\includegraphics[scale=0.35]{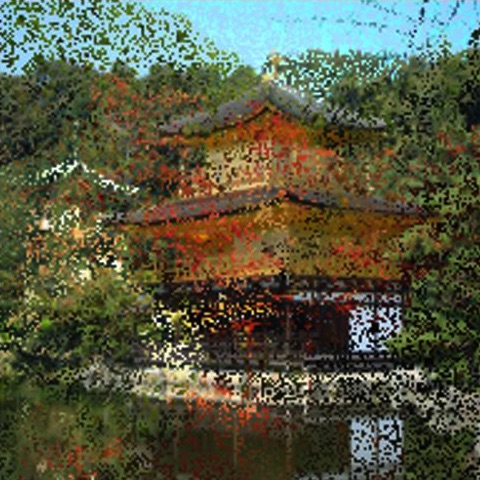} \hfil\\[-1.1\tabcolsep]
\vspace{0.3cm}
\includegraphics[scale=0.35]{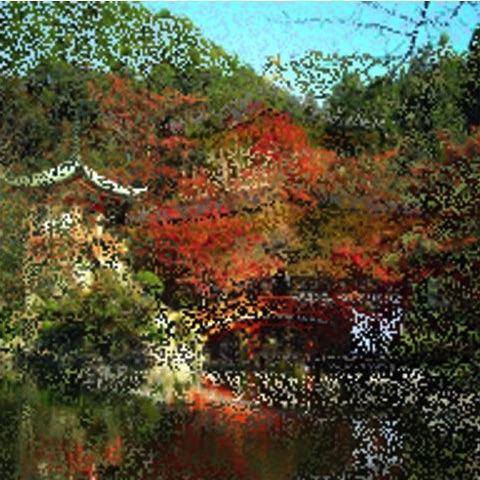}
\includegraphics[scale=0.337]{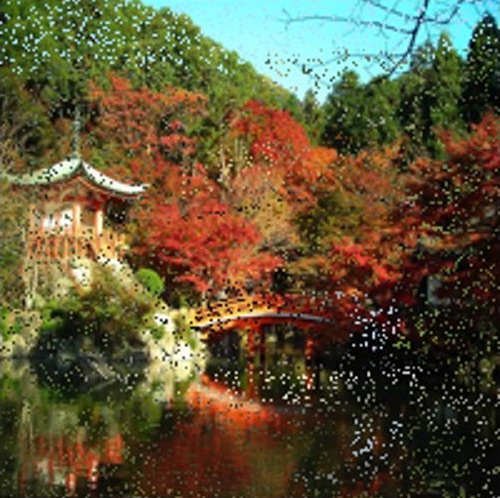} 

\vspace{0.3cm}
\caption{ Image transition for combined asymmetric and box mutation}

\label{fig:8}
\end{figure}

Interesting blurred elements randomly emerge over the image during the transition stage. Similar elements are shown as a characteristically steadily changing, whereas the elements of the target image continuously appear. On the last image we can almost completely see the details of the target image. The blurred elements seem to be created through the bright autumn colour in the target image in this stage of the sequence.

Figure~\ref{fig:7} shows the results of the experiments for \emph{combined asymmetric} and \emph{combined strip mutation}. 
The lower sequence of the picture is similar to the top in terms of the behaviour of the patches. More differences can be seen as the transition progresses, but the picture continues to be partially unrecognisable during the transition. New elements can be discovered on the affected image which slowly appear and create an element of surprise. 

We see less of the dotted image pattern, but now observe combined blurred elements randomly occurring in the upper sequence of the images. The image provides an interesting  comparison to the old linen pictures of the Rembrandt epoch.
In the advanced stages, the patches are well integrated into the images using \emph{asymmetric mutation}. Finally, the last image we see is well integrated and shows more detail in all areas of the image.  

Figure~\ref{fig:8} shows the results of the experiments for \emph{combined asymmetric} and \emph{box mutation}. 
In the advanced stages, the patches are well integrated into the images using \emph{asymmetric mutation} to produce a smooth transition process. In summary, the image transition of \emph{combined asymmetric} and \emph{box mutation} differs significantly from \emph{the box mutation} transition introduced in Section~\ref{sec4}.


\section{Conclusions and Future Work}
\label{sec6}

In this paper we have investigated how evolutionary algorithms and related random processes can be used for image transition.
We have shown how \emph{the asymmetric mutation} operator, introduced originally for the optimization of OneMax, can be applied to our problem. We examined how the varying mutation settings influences our results through \emph{box mutation, strip mutation, combined strip mutation} and combined approaches with \emph{asymmetric mutation}. By investigating combinations of the different approaches, we have shown a variety of interesting evolutionary image transition processes.
All our investigations are based on a fitness function that is equivalent to the well-known OneMax problem. It would be interesting to study more complex fitness functions and their impact on the artistic behaviour of evolutionary image transition in future research.

\bibliographystyle{abbrv}

\end{document}